\def\year{2019}\relax
\documentclass[letterpaper]{article}
\usepackage{aaai19}  
\usepackage{times}  
\usepackage{helvet}  
\usepackage{courier}  
\usepackage{url}  
\usepackage{graphicx}  
\usepackage[pdftex]{hyperref}
\usepackage{amsmath}
\usepackage{color}
\usepackage{textcomp}
\usepackage{booktabs}
\usepackage{ccicons}
\usepackage{todonotes}
\usepackage{amsfonts}
\usepackage{framed}

\newcommand{\cA}{\mathcal{A}}

\newcommand{\cN}{\mathcal{N}}
\newcommand{\cM}{\mathcal{M}}
\newcommand{\tit}[1]{\textit{#1}}
\newcommand{\IFF}{\textit{iff}}
\newcommand{\IF}{\textit{if}}
\newcommand{\ED}{\mathit{ED}}
\newcommand{\PCF}{\mathit{PCF}}
\newcommand{\walk}{\mathit{walk}}
\newcommand{\bike}{\mathit{bike}}
\newcommand{\car}{\mathit{car}}
\newcommand{\public}{\mathit{public}}
\newcommand{\taxi}{\mathit{taxi}}

\newcommand{\SUM}{\mathit{sum}}
\newcommand{\MAX}{\mathit{max}}
\newcommand{\MIN}{\mathit{min}}
\newcommand{\AVG}{\mathit{avg}}

\newcommand{\figref}[1]{Figure~\ref{fig:#1}}

\newcommand{\eqtref}[1]{Equation~\ref{eqt:#1}}

\begin{document}
%
\title{An Extensible and Personalizable Multi-Modal Trip Planner}
\author{
	Xudong Liu\\
	School of Computing\\
	University of North Florida\\
	Jacksonville, FL\\
	xudong.liu@unf.edu
	\And
	Christian Fritz\\
	Savioke Inc.\\
	San Jose, CA\\
	cfritz@savioke.com
	\And
	Matthew Klenk\\
	PARC, a Xerox Company\\
	Palo Alto, CA\\
	klenk@parc.com
}
\maketitle
\begin{abstract}
Despite a tremendous amount of work in the literature and in the commercial
sectors, current approaches to multi-modal trip planning still fail to consistently
generate plans that users deem optimal in practice.
We believe that this is due to the fact that current planners fail to capture
the true preferences of users, e.g., their preferences depend on aspects
that are not modeled. An example of this could be a preference not to walk
through an unsafe area at night.
We present a novel multi-modal trip planner that allows users to
upload auxiliary geographic data (e.g., crime rates) and to specify temporal
constraints and preferences over these data in combination with typical metrics
such as time and cost.
Concretely, our planner supports the modes walking, biking, driving, public
transit, and taxi, uses linear temporal logic to capture temporal constraints, and
preferential cost functions to represent preferences.
We show by examples that this allows the expression of very interesting
preferences and constraints that, naturally, lead to quite diverse optimal
plans.
\end{abstract}

\section{Introduction}
Trip planning, an application of planning and scheduling,
has seen substantive implementations by researchers and
developers \cite{bast2015route}.
Some of the planning systems are \tit{multi-modal}; that is,
combining distinct transportation modes, the trip planners
compute optimal routes from sources to destinations.
This notion of ``optimality" generally refers to the computed routes
having minimal total time or total fare.

However, in the eyes of a user, it may be more faceted than just
``fastest" or ``cheapest."
For instance, for a college student who specifies that she will
only walk or take public transit in a trip from Palo Alto to
San Francisco, the computed plan is not necessarily the fastest
(e.g., taking a cab could be faster)
or the cheapest (e.g., walking all the way has no fare).
This happens when a user tells the planner her hard constraints,
called \tit{constraints}.  
The planner then needs to either satisfy the 
constraints in the search process or return failure because they are
over-restrictive.
Moreover, the user might want to further customize the planner by
describing soft constraints, called \tit{preferences}.
For example, an agent wants to travel from school to
downtown, and prefers biking to taking the bus.
Thus, a trip with more biking than bus may be considered better for
the agent than the one with more bus than biking.
In this case, the planner will need to accommodate user preferences
whenever possible in the search of optimal solutions.

In the work by Yang et al. \cite{yang2009stated},
a pilot study was conducted to suggest favored transport
modes among the population in the Lisbon Metropolitan Area.
The study includes a survey involving 150 respondents, sampled
to roughly represent the socio-economic aspects of the local population.
Their results revealed that at least 72\% of the population picked
multiple travel modes (e.g., bus combined with heavy modes including subway,
train and ferry) over singular travel modes (e.g., private car, carpool and taxi).
The results also presented that almost half of the population had some
constraints on traveling time (e.g., departure times to/from work).
Furthermore, the pilot survey suggested correlations between
travel safety and travel modes, and between environmentally 
friendliness and travel modes.
To this end, our trip planning model is designed and developed in line
with these results.

Representing and reasoning about constraints and preferences are
fundamental to decision making in automated planning and scheduling
in artificial intelligence.
However, relatively limited effort has been devoted to designing and
implementing real-world multi-modal trip planners that captures user
constraints and preferences over the \tit{cost base}, possibly
extended from the user with auxiliary \tit{cost metrics},
such as crime rates and pollution statistics.
One notable work by Nina et al. \cite{nina2016no} introduced
system \tit{Autobahn} for generating scenic routes using
Google Street View images to train a deep neural network
to classify route segments based on their visual features.
Although \tit{Autobahn} computes scenic routes using computer
vision techniques, it does not account for extensibility
and personalizability.
Existing trip planners, such as SafePath \cite{galbrun2016urban}, 
and SocRoutes \cite{kim2014socroutes}, offer routes balancing between safety
and distance/time.
Nonetheless, these planners do not support personalization for the users.

Using a high-performance graph search
engine \cite{zhou2011dynamic}, we designed and implemented a 
multi-modal trip planner that uses pure
graph-search. This allows us to flexibly combine various modes
(i.e., walking, biking, driving, public transit, and taxi) and 
to declaratively specify constraints and preferences. 
The planner also allows the user
to upload \tit{new} mapping data over which
constraints and preferences can be expressed.  
For instance, a user might upload a map of crime
in the city, and ask the trip planner to avoid areas where crime is frequent.
To handle user constraints, the planner takes 
constraints (e.g., never bike after
transit, and never walk through bad neighborhoods) expressed in \tit{linear
temporal logic} to restrain the search space.
As with user preferences, the planner uses a
\tit{preferential cost function},  a weighted sum
over several cost metrics (e.g., time spend biking, fare on public transit, 
and overall crimes walking through) which can be re-weighted based
on different user preferences. 

Our paper is organized as follows.
In the next section, we present what it means for a planner to be extensible
and formally define the method to incorporating new metrics into the planner.
In the next section, we discuss the two aspects of personalization
in trip planning: constraints and preferences, and how they are represented and
reasoned with in the setting of multi-modal trip planning.
We then move on to describe the system structure of our graph-search based planner,
and show results obtained from our planner in various occasions.
We conclude by outlining future research directions.

\section{Extensibility}
Allowing users to upload their own data sets of interests is an important
step towards customization of a trip planner.
We designed a framework where a user can upload auxiliary cost 
metric data (e.g., crime
statistics and pollution data) into the planner,
and the planner will compute an optimal route accordingly.

The user-created data are the auxiliary data that is represented as pairs
of latitude and longitude degrees.
To merge these lat-long pairs into the planner, we performed a neighborhood
search to calculate the total score of auxiliary data for each 
lat-long pair already in our planner.
It might be of strong interest to some user for our planning
system to take care of criminal statistics so that some level of safety
of the resulting routes is guaranteed.
For instance, a user traveling through the downtown area of San Francisco
around midnight may want to upload a data set of crimes, 
and express her constraints and preferences in hope of a safer trip plan.
%

Formally, we denote by $\cA$ the set of auxiliary points
uploaded by the user, and $\cN$ the set of points in our planner.
Given a point $N=(x_N,y_N)\in \cN$ in our planner, an auxiliary
point $A=(x_A,y_A)\in \cA$ and an effective radius $r$,
we compute the auxiliary score $S(N,A,r)$ of $N$ contributed by $A$ 
with respect to $r$:
\[
	S(N,A,r)=
		\begin{cases}
			1 - \frac{\ED(N,A)}{r} & \text{if $\ED(N,A) \leq r$,}\\
			0 & \text{otherwise,}
		\end{cases}
\]
where $\ED(N,A)$ is the Euclidean distance between two points.
Thus, the auxiliary score $S(N,\cA,r)$ of $N$ for $\cA$ with respect to $r$
can be computed: $S(N,\cA,r)=\sum_{A \in \cA} S(N,A,r)$.

%

\section{Personalizability}
Personalizability consists of two aspects: constraints and preferences.
From the viewpoint of the planner,
constraints, also called hard constraints, are statements that the planner
has to satisfy during the planning process; whereas preferences, or
soft constraints, are specifications that the planner will need to optimize.
We formulated constraints using linear temporal logic (LTL) and preferences as
a preferential cost function (PCF), and implemented our planner leveraging the
widely-used graph search algorithm the A*.

\subsubsection{Constraints}
As constraints in the setting of trip planning are often declarative and
temporal, our choice of LTL is straightforward.
We now give a brief review of linear temporal logic (LTL).
Let $f$ be a propositional formula over a finite set $L$ of Boolean variables.  
LTL formulas are defined recursively as follows.
\begin{equation}
	\varphi = f | \varphi_1 \land \varphi_2 | \varphi_1 \lor \varphi_2 | \neg \varphi | 
		\bigcirc \varphi |	\Box \varphi | \Diamond \varphi | \varphi_1 \cA \varphi_2
\end{equation}
Note that we have $\varphi_1 \cA \varphi_2$, and it means that
``$\varphi_2$ holds right after $\varphi_1$ holds."

A natural constraint for an agent in trip planning could be ``In this trip 
I will not drive a car 
at all after biking or taking the public transit."
In LTL, such constraint can be translated into an LTL formula $\psi$
\begin{equation}
\label{eqt:ex}
	((M=\bike) \lor (M=\public)) \,\cA\, (\Box (\neg (M=\car))).
\end{equation}

Note that LTL allows agents to describe constraints over the entirety
of the search tree, not just limited to mode labels on edges.
For instance, an agent may also express ``In this trip I will 
bike for at least one hour but
not more than two," which in LTL would be
\begin{equation}
\label{eqt:ex2}
	(\Diamond (T_\bike \geq 3600)) \land (\Box (T_\bike \leq 7200)),
\end{equation}
where $T_\bike$ denotes the total time in seconds spent so far per $\bike$.

As the actions in trip planning is limited to taking different transportation modes,
in our definition of the semantics of LTL
these actions are subsumed into the interpretations of $L$, or \tit{states}.
The semantics of LTL is defined with regard to trajectories of states. 
Let $\sigma$ be a trajectory of states $S_0,a_1,S_1,\ldots,a_n,S_n$, and
$\sigma[i]$ a suffix $S_i, a_{i+1}, S_{i+1}, \ldots,a_n,S_n$.  We have
\begin{align*}
	\sigma \models f \;\; &\IFF \;\; S_0 \models f,\\
	\sigma \models \varphi_1 \land \varphi_2 \;\; &\IFF \;\; \sigma \models \varphi_1 \; and \; \sigma \models \varphi_2,\\
	\sigma \models \varphi_1 \lor \varphi_2 \;\; &\IFF \;\; \sigma \models \varphi_1 \; or \; \sigma \models \varphi_2,\\
	\sigma \models \neg \varphi \;\; &\IFF \;\; \sigma \not \models \varphi,\\
	\sigma \models \bigcirc \varphi \;\; &\IFF \;\; \sigma[1] \models \varphi,\\
	\sigma \models \Box \varphi \;\; &\IFF \;\; \forall 0 \leq i \leq n (\sigma[i] \models \varphi),\\
	\sigma \models \Diamond \varphi \;\; &\IFF \;\; \exists 0 \leq i \leq n (\sigma[i] \models \varphi),\\
	\sigma \models \varphi_1 \cA \varphi_2 \;\; &\IFF \;\; \forall 0 \leq i < n (\IF \; \sigma[i] \models \varphi_1, \sigma[i+1] \models \varphi_2).
\end{align*}

%

\subsubsection{Preferences}

A state is described as a set of \tit{state variables}.
The state variables of a state $S$ include the transportation mode $M$ that led to $S$,
time $T_M$ in seconds spent so far per mode $M$ (e.g., $T_\public$ for
public transit), fare $D_M$ spent so far per mode $M$ (e.g.,
$D_\taxi$ for taking a cab), and variables related to the auxiliary data once uploaded.
These extra data related variables are metrics such as the sum ($A_\SUM$),
the maximum ($A_\MAX)$, the minimum ($A_\MIN$), and the average ($A_\AVG$) data along the path.
We assume the fares $D_\walk$ and $D_\bike$ are zeros.

We denote by $\cM=\{\walk, \bike, \car, \public, \taxi\}$ the set of transportation
modes
and focus on weighted functions over state variables and
designed the cost function, called \tit{preferential cost function} (PCF), that guides the
graph-based search engine in our trip planner as follows.
\begin{equation}
	\begin{aligned}
		\PCF(S) = &\beta_T \cdot \sum\limits_{M \in \cM} (\alpha_M \cdot T_M)+
								 \sum\limits_{M \in \cM} D_M\\
							&	 +\beta_A \cdot A_\SUM,
	\end{aligned}
	\label{eqt:pcf}
\end{equation}
where $\alpha_M$ is the coefficient of $T_M$ specifying the relationship
between $M$ and $\car$,
and $\beta_T$ ($\beta_A$) is the ratio that describes how much 
in dollars a user would pay to
save an hour (an auxiliary datum, respectively).
Note that the PCF can be easily adjusted to cases when no auxiliary dataset
or multiple auxiliary datasets uploaded.

Clearly, to any given state our PCF assigns a monetary value,
the overall cost that drives our search algorithm in the planner.

\subsubsection{Preference Elicitation}
To gather these coefficients ($\alpha_i$'s and $\beta_i$'s) in our $\PCF$, we 
designed an interface to elicit them from the user.
The planner asks the user questions and collect answers from the user to derive the coefficients.
These questions are as follows.
\begin{enumerate}
	\setlength\itemsep{0pt}
	\item How many hours of driving do you think are equivalent to one hour of 
	      walking, biking, public transit, and taxi?
	\item How much in dollars would you pay to save an hour in traveling, and
	      to avoid an auxiliary datum (e.g., crime or pollution) in traveling?
\end{enumerate}
For instance, Alice, an agent, answers 3, 2, 0.25, 0.5, 20 and 1
to the questions above.
Intuitively, the numbers indicate that she prefers public transit the most, followed by
taxi, driving, biking and walking, in order.
We show how we can now derive $\alpha_i$'s in \eqtref{pcf}.
We start with setting $\alpha_\car=1$.
Now, since one hour of walking is equivalent to 3 hours of driving,
we have $\alpha_\walk \times 1 = \alpha_\car \times 3$;
hence, we derive $\alpha_\walk = 3$.
Similarly, we have $\alpha_\bike = 2$, $\alpha_\public = 0.25$,
and $\alpha_\taxi = 0.5$.
As with the other two coefficients $\beta_1$ and $\beta_1$,
we know one travel hour is worth 20 dollars and one auxiliary
event 1 dollar.  
We then have $\beta_1 \times 1 \text{ hour} = 1 \times 20 \text{ dollars}$
and $\beta_2 \times 1 \text{ aux} = 1 \times 1 \text{ dollars}$;
therefore, we derive $\beta_1=20\text{ dollars/hour}$ and 
$\beta_2=1\text{ dollar/aux}$.
Indeed, function $\PCF$ with the input of time, fare and
auxiliary metric pieces boils down to monetary cost,
and the planner computes the best path by optimize
based on this overall monetary cost in the searching
process.



Putting all together, we leveraged the widely-used A* search 
algorithm on top of our high-performance graph search
engine.  The A* algorithm incorporates the following cost function.
\begin{equation}
	f(S) = \PCF(S) + h(S),
\end{equation}
where $\PCF(S)$ is the overall cost of an optimal trip from the initial state to $S$, and
$h(S)$ is an admissible estimate of the cost of an optimal trip from $S$ to goal.
We set $h(S)$ the minimum estimate among all available modes in $S$.
To prune the search space, we check satisfiability of the temporal constraints in LTL
at expansion of the search tree.


\section{Implementation}

We designed and implemented a multi-modal trip planning
system 
based on a 
high-performance graph search engine.
The planner allows user uploads, as well as declarative
constraints and preferences.
We now describe the structure of the planning system.

The trip planner takes two types of data as input:
static data and user-specified request.
The static input includes Map Data and Transit
Data.
Map Data describes the map, a directed graph where 
nodes are street corners, bus stops and train stations.
Transit Data is a set of schedules for the buses and trains
On the other hand, a user provides her request, composed of 
three parts.
First, the user enters \tit{from} and \tit{to} locations 
on the map together with day and time of the start of the trip.  
Second, the user may upload her auxiliary metric dataset, e.g., crime rates.  
Lastly, the user specifies her constraints in LTL and preferences as a $\PCF$.
For example, the constraint 
could be ``never walk through a bad neighborhood."  Given these 
inputs, our planner computes an optimal path satisfying 
all the constraints and optimizing the preferences.


\section{Results}
First, we show the result computed from our planner for agent
Alice in the San Francisco Bay Area who commutes from Palo Alto to
San Francisco.
This is assuming no auxiliary metric datasets uploaded
so that the agent focus on time and fare.
Then, for agent Bob, we demonstrate the computed route in San Francisco
that is a much safer route than the quickest route.

\begin{figure}[!ht]
  \centering
    \includegraphics[width=0.45\textwidth]{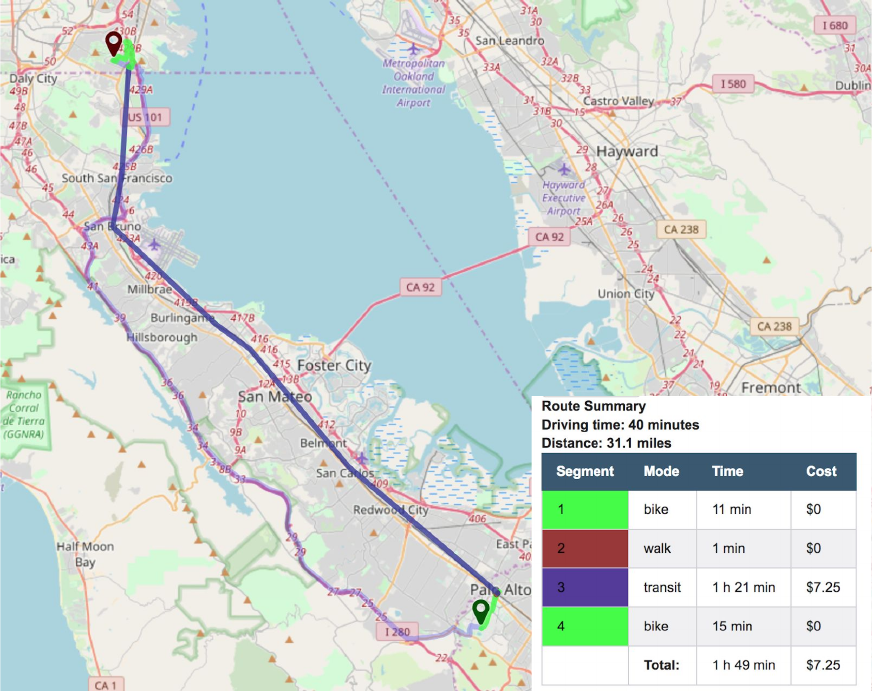}
  \caption{Resulting route for Alice\label{fig:alice}}
\end{figure}

Alice is constrained that she will not drive a car in her travel,
and her dollar per hour is thirty.
Moreover, Alice has a bicycle and
expresses that she will bike for at least 20 and at most 30 minutes.
So, Alice's constraint is specified as
$(\Box (\neg (M=\car))) \land (\Diamond (T_\bike \geq 1200)) \land (\Box (T_\bike \leq 1800))$.
Then, she expresses her preferences: biking and public transit are the most
preferred, next is taxi, and the least preferred is walking.
She has done so by answering the aforementioned elicitation questions,
and here we omit the detailed answers.
Note that the natural constraint in \eqtref{ex} 
is implicitly imposed on all cases, and that we consider
uberX for the taxi mode.
The result for Alice is depicted in \figref{alice}.
It spans 1 hour 49 minutes in time with the fare of 7 dollars 25 cents.
The path consists of 1 hour 21 minutes of pubic transit (i.e., Caltrain),
26 minutes of biking, and a minute of walking.
This is the optimal path satisfying Alice's constraints and prioritizing the
transit modes according to her preferences.

\begin{figure}[!ht]
  \centering
    \includegraphics[width=0.45\textwidth]{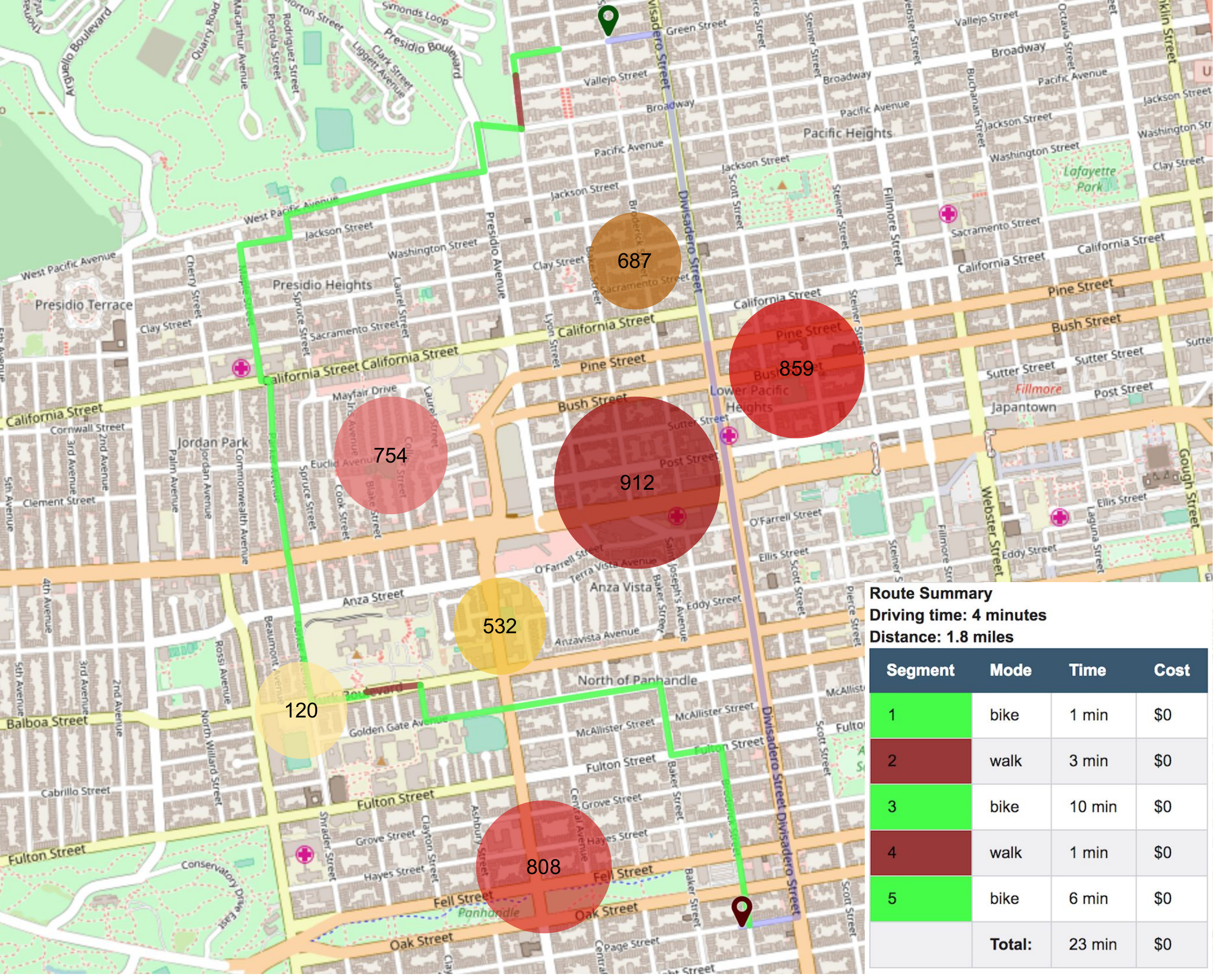}
  \caption{Optimal route considering crime rates for Bob\label{fig:crime}}
\end{figure}

%
Bob needs to travel via biking and walking across 
an area in northern San Francisco.
For Bob, safety is important.
Having found the crime statistics for the area, he uploads
the data as a new auxiliary metric into the map.
By specifying that he will never go through a neighborhood
with more than fifteen crimes over the last month, and that
he would sacrifice a quarter to avoid one crime incident,
the agent uses the planner to come up with a relatively safe
route. An example is shown in \figref{crime}, where several neighborhoods
are labeled by the crime counts.
The computed path is represented by the line colored by green and brown,
denoting biking and walking, respectively.
The quickest route is colored in light purple.
Clearly, this path routes away from crime-heavy areas and achieves
optimality in that the combined metrics -- time, fare and crime rates,
uploaded and personalized by the user -- is minimal among all possibilities.

\section{Future Work}
%
For the future, we plan to explore techniques in planning
for computing multiple routes that are diverse with bounded difference of costs.
Moreover, We intend to study the problem of learning the
$\PCF$ coefficients using the observations of the decisions the user made among
the computed paths. One possibility is to consider machine
learning algorithms such as linear and logistic regressions.
We also are interested in exploiting qualitative preference formalisms to be
embedded into trip planning, such as the well-known conditional preference
networks \cite{bbdh03} and 
lexicographic preference trees and forests \cite{booth:learningLP,conf/aaai15/LiuT,conf/foiks18/LiuT}.
Also interesting is to introduce traffic information into the planner
to support real-time multi-agent concurrent trip planning.

\bibliographystyle{aaai}
\bibliography{refs}

\end{document}